%% file: main.tex
\documentclass{ecai} 
\input{preamble}
\usepackage[capitalize,noabbrev]{cleveref}
\usepackage{tabularx} 

\begin{document}

\begin{frontmatter}
\paperid{1} 

\title{Evaluation of Black-Box XAI Approaches\\ for Predictors of Values of Boolean Formulae}
\author[A]{Stav Armoni-Friedmann}
\address[]{}
\author[A]{Hana Chockler}
\address[]{}
\author[A]{David A. Kelly}
\address[]{}


\address[A]{King's College London, UK}

\input{abstract}

\end{frontmatter}

\input{introduction.tex}
\input{related}
\input{cause}

\input{approach}

\input{algorithm}

\input{evaluation}

\input{conclusions}

\section*{Acknowledgments}
Hana Chockler and David A. Kelly acknowledge support of the UKRI AI programme and the Engineering and Physical Sciences Research Council 
for CHAI -- Causality in Healthcare AI Hub [grant number EP/Y028856/1].

\bibliography{all}


\end{document}

%% file: preamble.tex
\usepackage{amsmath}
\usepackage{amssymb}
\usepackage{amsthm}
\usepackage{booktabs}
\usepackage{graphicx}
\usepackage{color}
\usepackage{pifont}
\usepackage[utf8]{inputenc} 
\usepackage{booktabs}       
\usepackage{amsfonts}       
\usepackage{nicefrac}       
\usepackage{xcolor}         
\usepackage{subcaption}

\usepackage{tikz}
\usetikzlibrary{calc, positioning, shadows}
\usepackage{pgfplots}
\pgfplotsset{compat=1.18}

\usepackage{multirow}
\usepackage{algorithm}
\usepackage{algorithmic}
\usepackage{xspace}

\usepackage{microtype}
\usepackage{graphicx}
\usepackage{booktabs} 

\usepackage[pagebackref,breaklinks,colorlinks,allcolors=blue]{hyperref}

\newcommand{\rex}{\textsc{r}e\textsc{x}\xspace}
\newcommand{\xai}{XAI\xspace}
\newcommand{\gradcam}{\textsc{g}rad\textsc{c}am\xspace}
\newcommand{\lime}{\textsc{lime}\xspace}
\newcommand{\shap}{\textsc{shap}\xspace}

\newcommand{\anchors}{{\sc a}nchors\xspace}

\newcommand{\brex}{\textsc{b}-\textsc{r}e\textsc{x}\xspace}

\newtheorem{theorem}{Theorem}
\newtheorem{lemma}[theorem]{Lemma}

\newtheorem{proposition}[theorem]{Proposition}

\newtheorem{definition}{Definition}

\newcommand{\lem}{\begin{lemma}}
\newcommand{\elem}{\end{lemma}}
\newcommand{\pro}{\begin{proposition}}
\newcommand{\epro}{\end{proposition}}
\newcommand{\dfn}{\begin{definition}}
\newcommand{\edfn}{\end{definition}}
\newcommand{\K}{{\mathcal{K}}}

\newcommand{\bd}{\begin{definition}}
\newcommand{\ed}{\end{definition}}
\newcommand{\be}{\begin{enumerate}}
\newcommand{\bi}{\begin{itemize}}
\newcommand{\ee}{\end{enumerate}}
\newcommand{\ei}{\end{itemize}}

\newcommand{\U}{{\cal U}}

\newcommand{\cF}{{\cal F}}
\newcommand{\V}{{\cal V}}

\newcommand{\stam}[1]{}

\newcommand{\commentout}[1]{}

\newcommand{\clm}{\begin{claim}}
\newcommand{\eclm}{\end{claim}}

\newcommand{\thm}{\begin{theorem}}
\newcommand{\ethm}{\end{theorem}}

\definecolor{darkred}{rgb}{0.65,0,0}

\newcommand{\sat}{\models}

\newcommand{\xam}{\begin{example}}
\newcommand{\exam}{\end{example}}

\newcommand{\SWITCH}[1]{\STATE \textbf{switch} (#1)}
\newcommand{\ENDSWITCH}{\STATE \textbf{end switch}}
\newcommand{\CASE}[1]{\STATE \textbf{case} #1\textbf{:} \begin{ALC@g}}
\newcommand{\ENDCASE}{\end{ALC@g}}

\newcommand{\DEFAULT}{\STATE \textbf{default:} \begin{ALC@g}}
\newcommand{\ENDDEFAULT}{\end{ALC@g}}
\newcommand{\DEFAULTLINE}[1]{\STATE \textbf{default:} }

%% file: abstract.tex
\begin{abstract}
Evaluating explainable AI (XAI) approaches is a challenging task in general, due to the subjectivity of explanations. In this paper, we focus on tabular data and
the specific use case of AI models predicting the values of Boolean functions. We extend the previous work in this domain by proposing a formal and precise
measure of importance of variables based on actual causality, and we evaluate state-of-the-art XAI tools against this measure. We also present a novel XAI tool
\brex, based on the existing tool \rex, and demonstrate that it is superior to other black-box XAI tools on a large-scale benchmark. Specifically, \brex achieves
a Jensen-Shannon divergence of $0.072 \pm 0.012$ on random $10$-valued Boolean formulae.
\end{abstract}

%% file: introduction.tex
\section{Introduction}%
\label{sec:introduction}

AI models are now a primary building block of automated decision support systems. The opacity of some of these models (\emph{e.g.}, neural networks) creates demand for explainability techniques, which attempt to provide insight into why a particular input yields a particular observed output. Beyond increasing a user's confidence in the output, and hence also their trust in the AI model, these insights help to uncover subtle errors that are not detectable from the output alone~\cite{chockler2024}.

Despite a growing body of work in explainable AI (\xai), evaluating different approaches remains a difficult task due to often very different definition of explanation employed by different \xai tools. 
Current comparative methods, such as insertion/deletion curves~\cite{petsiuk2018rise}, sensitivity and infidelity~\cite{chih2019infidelity}, and user studies~\cite{SLKTF21} often work best on a particular subset of \xai approaches. Although these metrics are informative, they require principled inspection to draw conclusions and often fail to capture non-linear features of explanations. As a result of these challenges, evaluating explainability techniques remains an open problem.

\citet{EPOC2020} aim to bridge this gap using synthetic data which have an unambiguous `ground truth'; that is, there is a known (and unique) set of features per example which should be included in an explanation. Specifically, they design synthetic datasets that reflect truth tables of hand-crafted Boolean formulae. A feature is considered \emph{relevant} if it is decisive for the value of its parent node in a formula. For example, in the case of (True $\wedge$ False), it is the `False' which is \emph{relevant}, as changing its value changes the value of the entire formula. To test \xai tools using these definitions, they train a neural network to $100\%$ accuracy for each formula, and extract explanations, using various tools, for the outputs of these networks. They then report the proportion of times the result had a `perfect' overlap with the relevant features of each assignment. They only consider a small number of features, so perfection of overlap is limited to that number (the \emph{top-k} overlap). Reassuringly, most \xai tools performed perfectly on linear formula: less reassuringly, they did not work as well upon the introduction of non-linear operators.

Though informative, there remain a number of issues with the importance definition proposed, and methodology employed. Consider the case of (True $\lor$ True); here, either side of the truth an assignment can be changed without affecting the formula’s overall value. However, we argue that they \emph{both} remain important to the overall classification. Consider also a similar example (False $\lor$ False); here a change to \emph{either} assignment would change the formula’s overall value. We argue that all values in the above examples are relevant. 
Furthermore, the use of only a few hand-crafted formulae may introduce bias into results.

In this paper, we address the key limitations of the approach in~\cite{EPOC2020}, reproducing and extending their original experiment.
Instead of their binary definition of feature relevance, we use the notion of the \emph{degree of responsibility}~\cite{CH04}, which is grounded in the theory of actual causality~\cite{Hal19}. While they count \emph{top-k} overlap, we measure Jensen Shannon (JS-) divergence~\cite{JS91}: a symmetrical variant of the Kullback-Liebler divergence~\cite{kldivergence51} which penalizes mismatches, not just ranking mismatches. In addition, where they use a few hand-crafted formulae -- one linear, one nonlinear -- we use a large set of 
randomly generated formulae, making our
evaluation more reliable.

We present a linear-time algorithm for computing the precise degree of responsibility for read-once Boolean formulae. Unfortunately,
the exact computation is not as efficient in the general case, due to the intractability of the problem. We compute the ground-truth
responsibilities of features in the value of the formulae in our benchmark set, using the efficient algorithm for read-once formulae
and a brute-force exponential-time approach in the general case. The ground truth is used for evaluation.

We then present a novel black-box algorithm, \brex, based on the existing tool \rex~\cite{chockler2024}. 
\brex is similar to \rex in that it iteratively refines the search space to calculate approximate causal responsibility. 
However, the domain of application affects the algorithm: \rex is suited to explaining image classifiers, while
\brex explains tabular data classifiers, and specifically, Boolean formulae classifiers, which implies important differences
between the algorithms.

Our main contributions are as follows:
\begin{enumerate}
    \item An exact, ground-truth based, quality metric for evaluating XAI tools for Boolean formulae classifiers.
    \item An approximate black-box algorithm, \brex, based on the existing \rex tool.
    \item A quantitative comparison of popular XAI methods measuring JS-divergence against our exact algorithms; \brex performs
    significantly better than all other tools.
\end{enumerate}

The rest of the paper is organized as follows:~\Cref{sec:related} presents related work;~\Cref{sec:cause} reviews definitions of causal models and of responsibility;~\Cref{sec:approach} details algorithms derived from definitions of responsibility in Boolean formulae;~\Cref{sec:algo} presents \brex, and~\Cref{sec:evaluation} analyses the experimental results.

%% file: related.tex
\section{Related Work}\label{sec:related}

There is a large body of work on algorithms for computing an explanation
for a given output of a classifier. They can be largely grouped into white-box and black-box methods. In this context, white-box refers to methods which require access to the underlying model. Conversely, black-box methods interact exclusively with a models interface and observer it's behavior.
White-box methods frequently use variations on propagation-based
explanation methods to back-propagate a model's decision to the input
layer to determine the weight of each input
feature for the decision~\cite{springenberg2015striving,
sundararajan2017axiomatic, bach2015pixel, shrikumar2017learning,
nam2020relative}. \gradcam, a white-box technique which has spawned many variants, only needs
one backward pass and propagates the class-specific gradient into the final convolutional layer of a 
DNN to coarsely highlight important regions of an input image~\cite{CAM}. 

Perturbation-based approaches introduce perturbations to the input space. Perturbation methods are typically found in black-box explanation tools.
\shap (SHapley Additive exPlanations) 
computes Shapley values of different parts of the input and uses them to rank the
features of the input according to their importance~\cite{lundberg2017unified}.
\lime constructs a simple model to label the original input and its neighborhood of perturbed
images and uses this model to estimate the importance of different parts of the input
~\cite{lime}. \anchors uses a similar approach
to find parts of the inputs sufficient for the classification, regardless of the values
of other parts~\cite{RSG18}. 
Finally, \rex~\cite{chockler2024} ranks elements of the image according to their responsibility for the classification
and uses this ranking to greedily construct a minimal explanation sufficient for the original classification. 

There is also a growing body of work on logic-based explanations~\cite{INMS19,MSI22,DH23}, where a symbolic encoding
of the model is given. Their notion of abductive explanations is similar to the one used in this paper,
except all possible values of pixels outside of the explanations are considered. 
The problem setting is very different from ours, and they require that model usually be linear or monotonic. 
In contrast, our approach is black-box, is agnostic to the internal structure of the classifier, and does not make any assumptions about the nature of the classifier's decision process.

Rigorous, task-agnostic benchmarking of \xai approaches is still in its infancy, mostly due to the challenge of defining formal quality measures. The domain of Boolean functions is one of the rare examples where it is possible to 
accurately measure the quality of explanations, due to the presence of a ground truth. \citet{IRP19} compute causality
in Boolean formulae using SAT solvers. \citet{EPOC2020} measure efficacy by generating datasets for hand-crafted Boolean functions as a test bench. They use perfect top-$k$ overlap as a proxy for efficacy and conclude that even simple binary functions are hard to explain. We utilize their synthetic-data approach, but adopt a stricter measure for efficacy.

%% file: cause.tex
\begin{figure*}[t]
    \centering
    \begin{subfigure}{0.3\textwidth}
        \centering
        \begin{tikzpicture}[sibling distance = 2.5cm]
        \node  {$\land$}
            child {node {False};
             \node[red] {$\quad \quad \quad$ 1}
            }
            child {node {$\land$};
            \node[red] {$\quad \quad \quad$ 2}
                child {node {False};
                \node[red] {$\quad \quad \quad$ 1}}
                child {node {False};
                \node[red] {$\quad \quad \quad$ 1}}
           };
    \end{tikzpicture}
    \caption{Dependency values after executing \Cref{alg:dependencies}}\label{fig:example1}
    \end{subfigure}
    \hfill
    \begin{subfigure}{0.3\textwidth}
    \centering
     \begin{tikzpicture}[sibling distance = 2.5cm]
        \node {$\land$}
            child {node {False};
            \node[blue] {$\quad \quad \quad$ 2}}
            child {node {$\land$};
            \node[blue] {$\quad \quad \quad$ 1}
                child {node {False};
                \node[blue] {$\quad \quad \quad$ 2}}
                child {node {False};
                \node[blue] {$\quad \quad \quad$ 2}}};
    \end{tikzpicture}
     \caption{The values of $ctx$ during the execution of \Cref{alg:distribute}}\label{fig:example2}
    \end{subfigure}
    \hfill
    \begin{subfigure}{0.3\textwidth}
    \centering
    \begin{tikzpicture}[sibling distance = 2.5cm]
        \node  {$\land$}
            child {node {False};
             \node[orange] {$\quad \quad \quad \frac{1}{3}$}}
            child {node {$\land$}
                child {node {False};
                 \node[orange] {$\quad \quad \quad \frac{1}{3}$}}
                child {node {False};
                \node[orange] {$\quad \quad \quad \frac{1}{3}$}}
           };
    \end{tikzpicture}
    \caption{The final responsibility values assigned by \Cref{alg:distribute}}\label{fig:example3}
    \end{subfigure}
    \vspace{5mm}
    \caption{An example illustrating \Cref{alg:dependencies,alg:distribute}}\label{fig:example}
       \vspace{5mm}
\end{figure*}
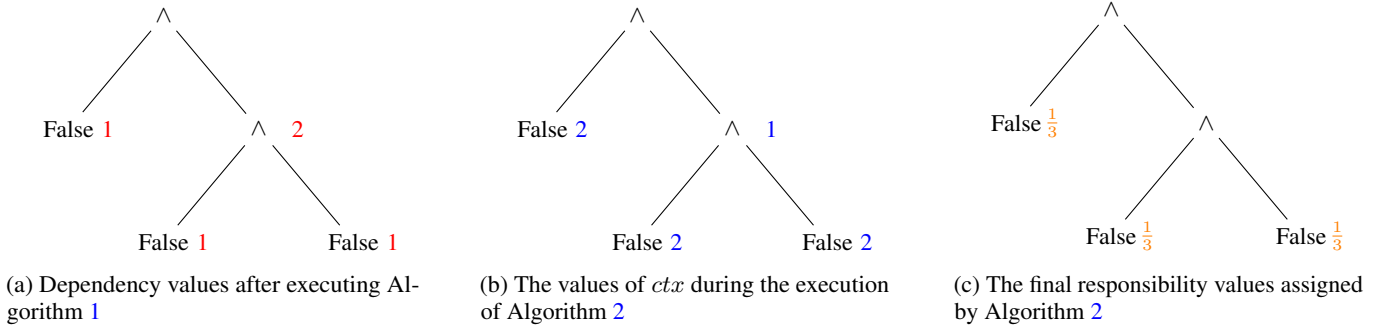

\section{Actual Causality}\label{sec:cause} 
In this section, we briefly review the definitions of causality and causal
models introduced by~\cite{HP01b} and
relevant definitions of causes and explanations in our setting by~\cite{CH24}. The reader is referred
to~\cite{Hal19} for further reading and definitions of causes and explanations in a general setting.

We assume that the world is described in terms of 
variables and their values.  
Some variables may have a causal influence on others. This
influence is modeled by a set of {\em structural equations}.
It is conceptually useful to split the variables into two
sets: the {\em exogenous\/} variables $\U$, whose values are
determined by 
factors outside the model, and the {\em endogenous\/} variables $\V$, whose values are ultimately determined by
the exogenous variables.  
The structural equations $\cF$ describe how these values are 
determined. A \emph{causal model} $M$ is described by its variables and the structural
equations. We restrict the discussion to acyclic (recursive) causal models.
A \emph{context} $\vec{u}$ is a setting for the exogenous variables $\U$, which then
determines the values of all other variables. 
We call a pair $(M,\vec{u})$ consisting of a causal model $M$ and a
context $\vec{u}$ a \emph{(causal) setting}.
An intervention is defined as setting the value of some
variable $X$ to $x$, and essentially amounts to replacing the equation for $X$
in $\cF$ by $X = x$. 
A \emph{probabilistic causal model} is a pair $(M,\Pr)$, where $\Pr$ is a probability distribution on contexts.

A causal formula $\psi$ is true or false in a setting.
We write $(M,\vec{u}) \sat \psi$  if
the causal formula $\psi$ is true in
the setting $(M,\vec{u})$.
Finally, 
$(M,\vec{u}) \sat [\vec{Y} \gets \vec{y}]\varphi$ if 
$(M_{\vec{Y} = \vec{y}},\vec{u}) \sat \varphi$,
where $M_{\vec{Y}\gets \vec{y}}$ is the causal model that is identical
to $M$, except that the 
variables in $\vec{Y}$ are set to $Y = y$
for each $Y \in \vec{Y}$ and its corresponding 
value $y \in \vec{y}$.

A standard use of causal models is to define \emph{actual causation}: that is, 
what it means for some particular event that occurred to cause 
another particular event. 
There have been a number of definitions of actual causation given
for acyclic models
(\emph{e.g.}, \cite{beckers21c,GW07,Hall07,HP01b,Hal19,hitchcock:99,Hitchcock07,Weslake11,Woodward03}).
In this paper, we focus on what has become known as the \emph{modified} 
Halpern--Pearl definition and some related definitions introduced
in~\cite{Hal19}. We briefly review the relevant definitions below.
The events that can be causes are arbitrary conjunctions of primitive
events (formulae of the form $X=x$). 

\dfn[Actual cause\label{def:AC}]
$\vec{X} = \vec{x}$ is 
an \emph{actual cause} of $\varphi$ in $(M,\vec{u})$ if the
following three conditions hold: 
\begin{description}
\item[{\rm AC1.}]\label{ac1} $(M,\vec{u}) \models (\vec{X} = \vec{x})$ and $(M,\vec{u}) \models \varphi$. 
\item[{\rm AC2.}] There is a
  a setting $\vec{x}'$ of the variables in $\vec{X}$, a 
(possibly empty)  set $\vec{W}$ of variables in $\V - \vec{X}'$,
and a setting $\vec{w}$ of the variables in $\vec{W}$
such that $(M,\vec{u}) \models \vec{W} = \vec{w}$ and
$(M,\vec{u}) \models [\vec{X} \gets \vec{x}', \vec{W} \gets
    \vec{w}]\neg{\varphi}$, and moreover
\item[{\rm AC3.}] \label{ac3}\index{AC3}  
  $\vec{X}$ is minimal; there is no strict subset $\vec{X}'$ of
  $\vec{X}$ such that $\vec{X}' = \vec{x}''$ can replace $\vec{X} =
  \vec{x}'$ in 
  AC2, where $\vec{x}''$ is the restriction of
$\vec{x}'$ to the variables in $\vec{X}'$.
\end{description}
\edfn
In the special case that $\vec{W} = \emptyset$, we get the 
but-for definition. A variable $x$ in an actual cause $\vec{X}$ is called a \emph{part of a cause}. In what follows, we adopt the
convention of Halpern and state that \emph{part of a cause is a cause}.

The notion of explanation taken from~\citet{Hal19} is relative to a set of contexts.
\dfn[Explanation\label{def:EX}]
$\vec{X} = \vec{x}$ is 
an \emph{explanation} of $\varphi$ relative to a set $\K$ of contexts 
in a causal model $M$ if the following conditions hold:  
\begin{description}
\item[{\rm EX1a.}]  
If $\vec{u} \in \K$ and $(M,\vec{u}) \models (\vec{X} = \vec{x})
  \wedge \varphi$, then there exists a conjunct $X=x$ of $\vec{X} =
  \vec{x}$ and a (possibly empty) conjunction $\vec{Y} = \vec{y}$ such
  that $X=x \wedge \vec{Y} = \vec{y}$ is an actual cause of $\varphi$
  in $(M,\vec{u})$. 
\item[{\rm EX1b.}] $(M,\vec{u}') \models [\vec{X} = \vec{x}]\varphi$  for all
  contexts $\vec{
    u}' \in \K$. 
\item[{\rm EX2.}] $\vec{X}$ is minimal; there is no
  strict subset $\vec{X}'$ of $\vec{X}$ such that $\vec{X}' =
  \vec{x}'$ satisfies EX1,  
where $\vec{x}'$ is the restriction of $\vec{x}$ to the variables in $\vec{X}'$. (This is SC4).
\item[{\rm EX3.}] \label{ex3} $(M,u) \sat \vec{X} = \vec{x} \wedge
  \varphi$ for some $u \in \K$.
\end{description}
\edfn

Following \cite{chockler2024}, we view a classifier (\emph{e.g.} a neural network)
as a probabilistic causal model. 
Specifically, the endogenous variables are taken to be the set $\vec{V}$ of inputs,
and the output variable $O$ is the output of the classifier. 
The equation for $O$ determines the output of the
model as an (unknown) function of the input values.
Thus, the causal model has depth $2$, with the feature variables determining
the output variable.  

We also assume \emph{causal independence} between the feature
variables $\vec{V}$. In the setting we adopt in our paper, $\vec{V}$ are the variables in
the Boolean formulae, and there is no dependency between them. 
We note that the causal independence assumption might not be true in
other types of inputs, such as tabular or spectral data. For those types of inputs, assuming independence
is clearly an approximation and might lead to inaccurate results.

Moreover, as the causal model is of depth $2$, all parents
of the output variable $O$ are contained in $\vec{V}$.  
Given these assumptions, the probability on contexts directly
corresponds to the probability on seeing various assignments to the Boolean variables.


Under the assumptions above, the following definition is equivalent to 
\Cref{def:EX}, as shown by~\cite{CH24}.

\begin{definition}[Explanation for neural networks]\label{defn:simple-exp}
An explanation is a minimal subset of variables of a
given input with their values that is sufficient for the model $\mathcal{N}$ to match the original
decision, with the values of other variables set to ``neutral'' values.
\end{definition}

In \brex algorithm, following the ideas introduced in~\cite{chockler2024},
the inputs are ranked according to their importance for the classification. Formally,
we consider \emph{singletons} (that is, a single variable and its value) that are \emph{parts of a cause}.
Recall that in Boolean formulae all variables are binary. The following definition, 
introduced in~\cite{chockler2024}, strictly speaking, defines what it means for a singleton to be a part of a cause
in neural networks.\footnote{The definition in~\cite{chockler2024} is restricted to image classifiers; we
generalize it here to all neural networks.} In the definitions below, we assume $(M,\vec{u})$.

\begin{definition}[Singleton cause in neural networks]\label{simple-cause}
For an input $\vec{V} = \vec{v}$ classified by the DNN as $O=o$, a variable $X \in \vec{X}$ and its value
$x \in \vec{x}$ is a \emph{singleton cause} (formally, a part of a cause)
of $O=o$ iff there exists a subset $\vec{W} \subset \vec{V}$ of variables such that the following conditions hold:
\begin{description}
 \itemsep0em
    \item[SC1.] $X \not\in \vec{W}$;
    \item[SC2a.] Changing the values of any subset $\vec{W}' \subseteq \vec{W}$ does not change $O$;
    \item[SC2b.] Changing the values of $\{X\} \cup \vec{W}$ changes $O$;
    \item[SC3.] $\vec{W}$ is a minimal subset of $\vec{V}$ that satisfies these conditions.
\end{description}
Such $\vec{W}$ is called a \emph{witness} for $X=x$ (note that $\vec{W}$ can be empty).
\end{definition}
\Cref{simple-cause} is equivalent to a part of a cause according to~\Cref{def:AC} in binary causal models
under the variable independence assumption~\cite{Bec22,chockler2024}. 

Finally, the following definition of the degree of responsibility quantifies causes as defined in~\Cref{simple-cause}.

\begin{definition}[Simplified responsibility]~\label{def:simple-resp}
The \emph{degree of responsibility} $r(X=x,O=o)$ of $X=x$ for $O=o$ is $1/(k+1)$, where $k$ is the size
of a smallest witness for $X=x$ being a singleton cause of $O=o$.
If $X=x$ is not a cause, we define $k = \infty$.
\end{definition}

%% file: approach.tex
\begin{figure*}[t]
\centering
  \includegraphics[width=0.8\linewidth]{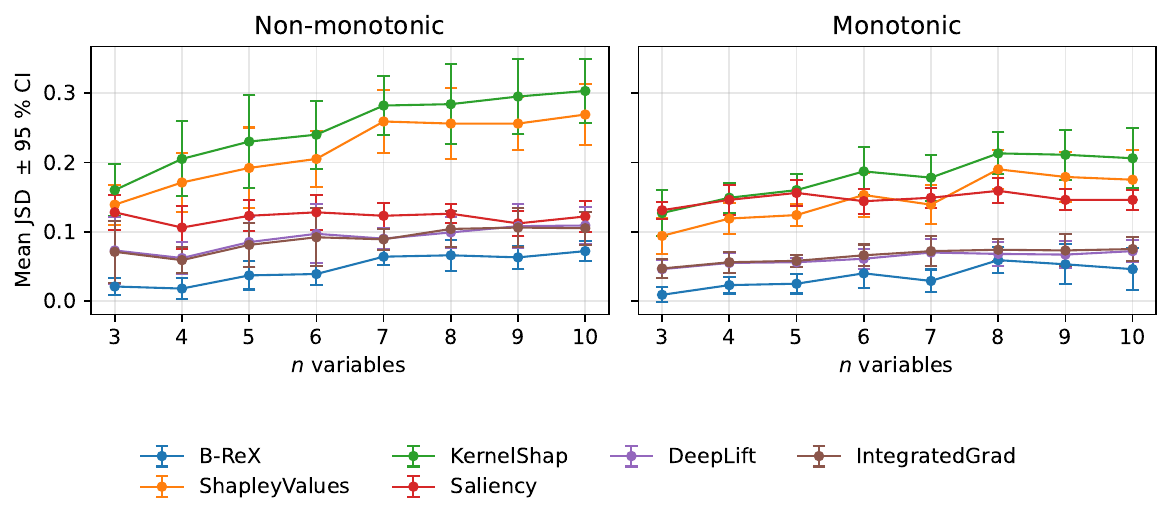}
  \caption{JSD from ground truth for monotonic and non-monotonic formulae}\label{fig:results}
  \vspace{5mm}
\end{figure*}

\section{Exact Computation for Boolean Formulae}\label{sec:approach}

In this section, we present exact algorithms for computing responsibility of assignments towards the overall truth value of formulae. We use these results as the ground truth responsibility values for the evaluation of XAI algorithms.

\subsection{Read-once formulae}
We first present a linear-time algorithm for read-once Boolean formulae. A read-once formula is a boolean expression where each variable occurs only once. Computing responsibility in tree-structured causal graphs is tractable and can be computed in time linear to the number of edges in the tree: $O(|E|)$~\cite{EL06}.

Given a Boolean formula $\varphi$, we construct a causal model, $M_\varphi$, by utilizing the tree structure of the formula.
Specifically, given $\varphi(x_1,\dots,x_n)$, with internal gates ${g_1 \dots g_m}$, and $g_m$ as the output node. Each gate reflects the value of some subtree of $\varphi$. Then we construct the model as follows:

\begin{enumerate}
\item The set of endogenous variables $\V$ is defined as 
\[ \V = \{x_1, \ldots, x_n \} \cup \{ g_1, \ldots, g_m \} \cup \{ O \}, \] 
where $\{x_1, \ldots, x_n \}$ is the set of variables of $\varphi$, $\{ g_1, \ldots, g_m \}$ 
is the set of gates of $\varphi$, and $O$ is a single output node,
whose value is equal to the value of $\varphi$.
\item For each gate (a unary or a binary operator) in the parse tree of $\varphi$, the structural equation defining the value 
of $g_i$ of the form $g_i \gets g_l \circ g_r$ for a binary operator $\circ$ and descendants $g_l$ and $g_r$, and
is $\neg g_l$ for the unary operator $\neg$. 
\item All variables are binary.
\end{enumerate}

We compute responsibility in two passes. In the first pass (\Cref{alg:dependencies}), we count, for each endogenous variable, the minimum number of leaf node interventions required to change its value. In the second pass (\Cref{alg:distribute}), we use this information to calculate the degree of responsibility at each leaf node.

We introduce a function $\mathcal{I}$, which, for any node in the causal model $M_\varphi$, indicates which subtrees are causes of the current value. Formally, for an internal node $g_i$ with children $g_l$ and $g_r$ $g_i = g_l \circ g_r, \; \circ \in \{\wedge, \lor, \oplus \}$ or $g_i = \neg g_l$, where the value of $\mathcal{I}(g_i)$ is
\[
    \mathcal{I}(g_i) =
    \begin{cases}
        \text{Both} & \neg g_l \circ \neg g_r \neq \varphi \\
        \text{Either} & \neg g_l \circ g_r \neq \varphi \wedge g_l \circ \neg g_r \neq \varphi \\
        \text{Left} & \neg g_l \circ g_r \neq \varphi \\
        \text{Right} & g_l \circ \neg g_r \neq \varphi \\
        \text{Pass} & \neg g_l \\
    \end{cases}
\]

Intuitively, the result of the $\mathcal{I}$ operator denotes the cause for the value of the variable $g_i$ in terms of the values of
its direct descendants.

\Cref{alg:dependencies} describes a single-pass algorithm which labels each variable in $M_\varphi$ that corresponds to
a gate of $\varphi$ (that is, not an input) with the information required to calculate exact causal responsibility in the second pass over the model (\Cref{alg:distribute}).

\begin{algorithm}[h]              
\caption{DEPENDS($g_i$)}\label{alg:dependencies}            
\begin{algorithmic}[1]
  \REQUIRE $deps: g_i \rightarrow \mathbb{Z}$ (default 0)
  \IF{$g_i \text{ is a \textbf{leaf}}$}
    \STATE $deps(g_i) \leftarrow 1$
  \ELSE
    \STATE $d_j, d_k \gets \textsc{depends}(g_l), \textsc{depends}(g_r)$
    \STATE deps$(g_i) \gets
    \begin{cases}
        d_j + d_k  & \mathcal{I}(g_i) = \textbf{Both} \\
        min\{d_j, d_k\} & \mathcal{I}(g_i)  = \textbf{Either} \\
        d_j & \mathcal{I}(g_i)  = \textbf{Left} \\
        d_k & \mathcal{I}(g_i)  = \textbf{Right} \\
        d_j & \mathcal{I}(g_i)  = \textbf{Pass}
    \end{cases}$
  \ENDIF
\end{algorithmic}
\end{algorithm}
\noindent

\begin{lemma}
\label{lem:deps}
DEPENDS($O$) is the smallest number of variables in $\varphi$ which need to change value in order to change the value of $\varphi$.
\end{lemma}
\noindent
The proof is by induction on the structure of the formula $\varphi$.

~\Cref{alg:distribute} uses the output of ~\Cref{alg:dependencies} to compute the degrees of responsibility of each variable in the value of $\varphi$. During the top-down traversal, we maintain an accumulator $ctx$ equal to the size of the witness set collected so far. If a parent node depends on only one descendant, we pass $ctx$ unchanged to that branch. If it depends on both descendants, each branch inherits $ctx$ plus the cheapest flip cost of its sibling ($deps[sib]$). When the recursion reaches a leaf, we simply set its responsibility value to $\frac{1}{1+ctx}$.

\Cref{alg:distribute} propagates the values returned by \Cref{alg:dependencies} from the root node to its leaves to calculate responsibility. At each point in the recursion $ctx$ stores the number of leaf interventions required to make the value of the $O$ counterfactually dependent on the current node.

\begin{algorithm}[h]
\caption{RESPONSIBILITY($g_i, ctx$)}
\label{alg:distribute}            
\begin{algorithmic}[1]
  \REQUIRE $deps: g_i \rightarrow \mathbb{Z}$
  \REQUIRE $\rho: \mathcal V \rightarrow [0, 1]$
  \IF{$g_i \text{ is a} \textbf{ leaf }$}
    \STATE $\rho[g_i] \leftarrow 1/(1 +ctx)$
  \ELSE
    \SWITCH{$\mathcal{I}(g_i)$}
        \CASE{\textbf{Left} or \textbf{Pass}}
            \STATE \textsc{responsibility$(g_l, ctx)$}
        \ENDCASE
        \CASE{\textbf{Right} }
            \STATE \textsc{responsibility$(g_r, ctx)$}
        \ENDCASE
        \CASE{\textbf{Either}}
          \STATE \textsc{responsibility$(g_l, ctx)$}
           \STATE  \textsc{responsibility$(g_r, ctx)$}
           \ENDCASE
        \CASE{\textbf{Both}}
            \STATE  \textsc{responsibility$(g_l, ctx + deps(g_r))$}
            \STATE  \textsc{responsibility$(g_r, ctx + deps(g_l))$}
        \ENDCASE
    \ENDSWITCH
  \ENDIF
\end{algorithmic}
\end{algorithm}

Both \Cref{alg:dependencies} and \Cref{alg:distribute} run in time linear in the size of $\varphi$, as they perform $O(|\varphi|)$
operations, each of which takes constant time. 

\begin{table*}[t]
    \caption{Accuracy with respect to $top$-$k$ features on non-monotonic formulae}
    \label{table:legacy_nonlinear}
    \centering
    \begin{tabular}{l|cccccccc}
        \hline
        Models & 3 & 4 & 5 & 6 & 7 & 8 & 9 & 10 \\
        \hline
        \brex 
 & \textbf{0.754} & \textbf{0.782} &	\textbf{0.582} &	\textbf{0.398} &	\textbf{0.344} &	\textbf{0.368} &	\textbf{0.426}	& \textbf{0.384} \\
ShapleyValues&	0.404&	0.266&	0.296&	0.152&	0.262&	0.162 &	0.32&	0.15\\
KernelShap&	0.03&	0.002&	0.038&	0.002&	0.034&	0.004&	0.038&	0.004\\
Saliency&	0.23&	0.238&	0.374&	0.268&	0.334&	0.262&	0.4	&0.242\\
DeepLift&	0.25&	0.256&	0.264&	0.182&	0.256&	0.268	&0.31&	0.122\\
InputXGrad &	0.002&	0&	0&	0&	0	&0	&0	&0\\
\hline
    \end{tabular}
\end{table*}

\begin{table*}[t]
    \caption{Accuracy with respect to $top$-$k$ features on monotonic formulae}
    \label{table:legacy_linear}
    \centering
    \begin{tabular}{l|cccccccc}
        \hline
        Models & 3 & 4 & 5 & 6 & 7 & 8 & 9 & 10 \\
        \hline
        \brex            & \textbf{0.774} & \textbf{0.692} & \textbf{0.630} & \textbf{0.648} & \textbf{0.662} & \textbf{0.676} & \textbf{0.642} & \textbf{0.664} \\
        ShapleyValues  & 0.704 & 0.512 & 0.506 & 0.474 & 0.464 & 0.576 & 0.518 & 0.486 \\
        KernelShap     & 0.050 & 0.032 & 0.042 & 0.036 & 0.044 & 0.050 & 0.042 & 0.042 \\
        Saliency       & 0.458 & 0.432 & 0.416 & 0.460 & 0.342 & 0.538 & 0.334 & 0.382 \\
        DeepLift       & 0.650 & 0.524 & 0.538 & 0.520 & 0.458 & 0.650 & 0.470 & 0.472 \\
        InputXGrad     & 0.446 & 0.008 & 0.382 & 0.048 & 0.486 & 0.088 & 0.334 & 0.032 \\
        \hline
    \end{tabular}
    \vspace{5mm}
\end{table*}

\begin{theorem}\label{thm:resp}
For a read-once formula $\varphi$, for each variable $x_i$, \textsc{responsibility}$(x_i)$ returns the degree of responsibility,
defined in \Cref{def:simple-resp},
of the value of $x_i$ for the value of the output in $M_\varphi$.
\end{theorem}
\noindent
The proof is, again, by induction on the structure of $\varphi$.

~\Cref{fig:example} Shows the execution of ~\Cref{alg:dependencies,alg:distribute} on the formula $(False \land (False \land False))$

\subsection{General formulae}

In the general case, the problem of computing the degree of responsibility is intractable and hence can not be computed precisely
in a constant number of passes on the model (or in polynomial time). 
Therefore, for general formulae we instead employ the brute-force approach
of enumerating the full truth table for $\varphi$. The brute-force approach is exponential in the size of the formula and is thus applicable only to small formulae.

%% file: algorithm.tex
\section{The \brex Algorithm}\label{sec:algo}

In this section we present the \brex algorithm. \brex is completely model agnostic and ranks the variables according to their
degree of responsibility in the model's classification using the model's output only. 
The approach follows \rex in that it recursively partitions the input space in search of causes for which we compute (approximate) responsibility. 

We begin by partitioning our input into disjoint sets, such that for an input vector $X$ of length $n$, and a parameter $b$ for the number of parts, we partition
the indices $1, \ldots, n$ to $b$ sets of indices $\mathcal{B}: \{P_1\dots P_b\}$. A \emph{mutant} $X^{\sim P_i}$ is an input that has been modified by 
masking (setting to `Unassigned') the values of variables with indices not in $P_i$. We now apply our earlier definitions of responsibility, considering each partition to be a single point for intervention. We then observe the combinations of interventions that result in counterfactual dependencies, and distribute responsibility accordingly.

Specifically, we say that the degree of responsibility of a block P is the size of the smallest combination of blocks which cause our classifier to become counterfactually dependent on $P$. It follows that for a part $P \in \mathcal{B}$ the following expression gives the smallest such set of blocks:
$$
W^\star = \min_{W\in \mathcal{P}(\mathcal{B})}
\left\{
\begin{aligned}
|\mathcal B \setminus W|: & \varphi(X^{\sim W}) & \neq \varphi(X) \\
      & \varphi(X^{\sim(P \cup W)})&=\varphi(X)
\end{aligned}
\right\}
$$
i.e.\ the size of the \emph{smallest} set of \emph{other}
blocks that renders the classifier counterfactually dependent on
$P$. For each index in P, we then set its responsibility value to $\frac{1}{1+|W^\star|}$. It is also important to note that there may not be such a combination of blocks, in which case we set each value in P to 0, reflecting the fact that it does not satisfy the definition for a cause.

Next, we recurse on each partition for which $W^\star$ exists. The recursive pass conducts the same procedure, refining responsibility estimates within the subject part. The \brex algorithm for computing an approximate responsibility is presented in \Cref{alg:brex}.

\begin{algorithm}[H]
\caption{\brex$(P)$}
\label{alg:brex}
\begin{algorithmic}[1]
\REQUIRE $\rho$: global responsibility map (vector length $n$)
\REQUIRE $P$: current index set to refine
\STATE $\mathcal B \leftarrow \textsc{Partition}(P,\rho)$
\FOR{$P_i \in \mathcal{B}$}
  \IF{$W^\star$ is defined for $P_i$} 
    \STATE $\rho[P_i] \gets \dfrac{1}{1 + |W^*|}$
    \STATE $\operatorname{\brex}(P_i)$
  \ELSE
    \STATE $\rho[P_i] \gets 0$
  \ENDIF
\ENDFOR
\end{algorithmic}
\end{algorithm}

Here, we note that the partitioning strategy is not random. Rather, the algorithm uses previously computed responsibility maps to direct partitioning. Specifically, the partitioning strategy attempts to distribute indices such that the sum of responsibility values for each partition is roughly equal. Practically, this means that areas of low responsibility in the input are grouped, while more pertinent areas of the input are broken up and investigated more closely.

The partitioning strategy employed simply takes the $l_1$ norm of the responsibility vector to use as weights. Indices are probabilistically selected for addition to a partition until the cumulative weight of the partition exceeds $1/m$ where $m$ is the number of partitions requested.

%% file: evaluation.tex
\begin{table*}[ht]
    \caption{JSD from ground truth on non-monotonic formulae}%
    \label{tab:new_nonlinear}
    \centering
    \begin{tabular}{l|cccccccc}    
        \hline
        & \multicolumn{8}{c}{Formula arity} \\
        & 3 & 4 & 5 & 6 & 7 & 8 & 9 & 10\\
        \hline
        \brex & \textbf{0.021} & \textbf{0.018} & \textbf{0.037} & \textbf{0.039} & \textbf{0.064} & \textbf{0.066} & \textbf{0.063} & \textbf{0.072} \\
        ShapleyValues & 0.139 & 0.171 & 0.192 & 0.205 & 0.259 & 0.256 & 0.256 & 0.269 \\
        KernelShap & 0.160 & 0.205 & 0.230 & 0.240 & 0.282 & 0.284 & 0.295 & 0.303 \\
   
        Saliency & 0.128 & 0.106 & 0.123 & 0.128 & 0.123 & 0.126 & 0.112 & 0.122 \\
        DeepLift & 0.073 & 0.062 & 0.085 & 0.097 & 0.090 & 0.099 & 0.108 & 0.109 \\
        InputXGrad & 0.319 & 0.301 & 0.298 & 0.400 & 0.374 & 0.279 & 0.382 & 0.342 \\
        IntegratedGrad & 0.071 & 0.059 & 0.081 & 0.092 & 0.089 & 0.104 & 0.106 & 0.105 \\
        \hline
    \end{tabular}
\end{table*}

\begin{table*}[ht]
    \caption{JSD from ground truth on monotonic formulae}%
    \label{tab:new_linear}
    \centering
    \begin{tabular}{l|cccccccc}
        \hline
        & \multicolumn{8}{c}{Formula arity} \\
        & 3 & 4 & 5 & 6 & 7 & 8 & 9 & 10\\
        \hline
\brex &	\textbf{0.009} &	\textbf{0.023}&	\textbf{0.025}&	\textbf{0.040} &	\textbf{0.029} &	\textbf{0.059}&	\textbf{0.053}&	\textbf{0.046}\\
ShapleyValues	&0.094&	0.119&	0.124&	0.153&	0.139&	0.190&	0.179&	0.175\\
KernelShap&	0.127&	0.149&	0.160&	0.187&	0.178&	0.213	&0.211&	0.206\\
Saliency&	0.131&	0.146&	0.156&	0.144&	0.149&	0.159&	0.146&	0.146\\
DeepLift&	0.046&	0.055&	0.056&	0.061&	0.070&	0.068&	0.067&	0.072\\
InputXGrad&	0.499&	0.392&	0.362&	0.393&	0.407&	0.339&	0.399&	0.487\\
IntegratedGrad&	0.047&	0.056&	0.058&	0.066&	0.072&	0.074&	0.073&	0.075\\
\hline
\end{tabular}
\vspace{5mm}
\end{table*}

\section{Evaluation}
\label{sec:evaluation}

We measured how closely attributions returned by B-ReX and other popular algorithms match the ground-truth responsibility maps calculated
in \Cref{sec:approach}. We evaluated each explainer against randomly constructed Boolean formulae and reported their Jensen-Shannon divergence(JSD)~\cite{JS91} from the ground truth. In order to test gradient-based methods, we trained MLP networks until 0\% loss on each randomly constructed formula, and run white box explainers (IntegratedGradients, DeepLift, Saliency) on these.

\subsection{Experimental Setup}

We tested explainers from the Captum \cite{captum} library at default settings, and our own implementation of \brex. 
We generated $10$ Boolean expressions of $3$-$10$ variables. 
For generation of monotonic formulae, we restricted the operators to logical AND and OR. For generation of non-monotonic formulae, we also allowed NOT and XOR. 
We used strong 3-valued Kleene logic~\cite{K3} (K3) for the semantics of three-valued Boolean formulas, allowing the variables to have values True, False, or Unassigned.

For each generated formula, we trained a fully connected feed-forward neural network with ReLu activation to 100\% accuracy. Each network has six $20$-node layers, with a $12$-node input layer. Each explainer was executed on a model for each random formula for the entire set of possible assignments. 

The experiment used input vectors of length $12$, where the formulae had $3$-$10$ variables. $10$ such formulae were generated, and the explainers were evaluated against the full set of assignments for all $12$ variables. The choice of 12 variables was made to maintain comparability with experiments by Trischer et al. and to ensure reasonably sized truth-tables. For each returned attribution map, we measured the Jensen-Shannon Divergence  (JSD) to the ground-truth values, computed either by the brute-force approach or, for read-once formulae, by~\Cref{alg:distribute}. Here, JSD is used as it handles zero values in a principled manner.
\Cref{fig:results} shows the results of these tests, with error bars at the 95\% confidence interval.

\subsection{Results}
We first reproduced results from \cite{EPOC2020}, with the inclusion of \brex, and we show the results in 
\Cref{table:legacy_small,table:legacy_linear,table:legacy_nonlinear}.
\begin{table}[h]
    \vspace{2mm}
    \caption{Accuracy with respect to the top $k$ features}
    \label{table:legacy_small}
    \centering
    \begin{tabular}{lcccc}
        \hline
        Formulae & $\land$ & $\lor$ & $\oplus$ &$\oplus \land \oplus$\\
        \hline
        \textbf{\brex} & \textbf{0.992} & \textbf{0.986} & \textbf{1} & 0.638 \\
        ShapleyValues & 0.88 & 0.878 & 0.692 & 0.536 \\
        KernelShap & 0.068 & 0.054 & 0.024 & 0.006 \\
        Saliency & 0.842 & 0.742 & 0.684 & \textbf{0.774} \\
        DeepLift & 0.958 & 0.806 & 0.826 & 0.734 \\
        InputXGrad & 0.136 & 0.642 & 0.198 & 0.314 \\
        \hline
    \end{tabular}
\end{table}

\Cref{table:legacy_small} shows the proportion of times that the top features of an attribution map coincided perfectly with the $relevant$ features of an assignment. For each formula, exhaustive this proportion was aggregated over the entire set of truth table assignments. For example, for the case of $\land$, the input to the model would be $\{x_1, \dots, x_{12}\}$, where the function learned by the model is $(x_1 \oplus x_2) \land (x_3 \oplus x_4)$. Here we see that though ReX outperforms models in almost all cases, all methods struggle to explain non-monotonic formulas.

\Cref{table:legacy_linear} measures the same metric, but for a linear Boolean formulae of a fixed structure $(((x_1 \land x_2) \lor x_3) \land \dots )$ and varying size. Here performance degrades gracefully with formula size for all explainers.

\Cref{table:legacy_nonlinear} displays information similar to \Cref{table:legacy_linear} but for nonlinear formulae. However, the formula being tested is of the form $(((x_1 \oplus x_2) \land x_3) \oplus \dots)$. The non-monotonicity of this function leads to significantly degraded performance compared with monotonic formulae.

The results on the benchmark set consisting of randomly generated formulae are presented in~\Cref{tab:new_linear,tab:new_nonlinear}, which also shows the average average JS-divergence between attribution maps returned by each explainer on $10$ randomly constructed formulae of each arity. The same information is displayed graphically in \Cref{fig:results}, with error bars at the 95\% confidence interval. Results here agree with the reproduced work. It is also clear that while other black-box methods significantly degrade for non-monotonic formulae, \brex maintains superior performance.

\subsection{Discussion}
\brex consistently delivers the lowest JS divergence for each formula size, and on both formula families. For non-linear cases, \brex scales better than other black-box explainers (KernelShap, ShapleyValues), but not as well as gradient-based methods.

Both KernelShap and ShapleyValues are derived from co-operative game theory, and hence attempt to approximate the $additive$ contribution of each input. In the monotonic case, this is a valid assumption, however, when testing in non-monotonic settings, such
as formulas with XOR, this assumption becomes problematic. Consider the case of a 'True' value in the left-hand side of an XOR; it may increase or decrease the output in a context-dependent manner. This results in Shapley esimates being highly sensitive to the sampling strategy, hence negatively affecting performance.

Gradient-based explainers, such as Saliency, Integrated Gradients, or DeepLift, treat the gradient of $\nabla_{x_i}\varphi(\mathbf x)$ as a proxy
for feature importance.  This strategy is effective when $\varphi$ is differentiable in the neighborhood of $\mathbf x$, 
but breaks down when this is not the case (for example, with XOR). Our trained classifiers predict each boolean function with $100\%$ accuracy: even though XOR is not differentiable, the model has still learned to compute a gradient at this point. This gradient likely contains some interpolation learned from other features which are differentiable and may not be a completely faithful representation of the underlying Boolean function. Hence, the gradient signal spreads relevance across multiple inputs, lowering JSD compared with SHAP’s sampling bias, but still
over-estimating features that are \emph{not} counterfactually decisive.

\brex does not assume differentiability or monotonicity of the Boolean formula, leading to better performance in non-monotonic settings. Additionally, \brex scales better in both monotonic and non-monotonic cases, compared to other black-box explainers. However, \brex is slower than gradient-based methods, which is to be expected, as it needs multiple forward passes as opposed to a single backward pass.

\subsection{Limitations}
This study is deliberately limited to classification of values of Boolean formulae as these translate in a natural way to 
causal models and allow for
a well-defined ground truth. Though this provides a clean benchmark, it also limits generality. 
Furthermore, the precise computation of the ground truth is intractable, forcing us to use a brute-force
exponential computation, thus our approach does not scale well to formulae with a large number of variables.

%% file: conclusions.tex
\section{Conclusions}
\label{sec:conclusions}
We proposed an evaluation setting for \xai tools by adopting the degree of causal responsibility as a gold-standard explanation
measure. We suggested a linear-time ground-truth computation for read-once formulas and a model-agnostic (black-box) algorithm
\brex for explaining the decisions of classifiers on Boolean formulae. Our experimental results on a benchmark of randomly generated
Boolean formulae demonstrate the superior accuracy of \brex compared to other, both black-box and white-box, explainers.


%% file: main.bbl
\begin{thebibliography}{35}
\providecommand{\natexlab}[1]{#1}
\providecommand{\url}[1]{\texttt{#1}}
\expandafter\ifx\csname urlstyle\endcsname\relax
  \providecommand{\doi}[1]{doi: #1}\else
  \providecommand{\doi}{doi: \begingroup \urlstyle{rm}\Url}\fi

\bibitem[Bach et~al.(2015)Bach, Binder, Montavon, Klauschen, M{\"u}ller, and Samek]{bach2015pixel}
S.~Bach, A.~Binder, G.~Montavon, F.~Klauschen, K.-R. M{\"u}ller, and W.~Samek.
\newblock On pixel-wise explanations for non-linear classifier decisions by layer-wise relevance propagation.
\newblock \emph{PLOS One}, 10\penalty0 (7), 2015.

\bibitem[Beckers(2021)]{beckers21c}
S.~Beckers.
\newblock Causal sufficiency and actual causation.
\newblock \emph{Journal of Philosophical Logic}, 50:\penalty0 1341--1374, 2021.

\bibitem[Beckers(2022)]{Bec22}
S.~Beckers.
\newblock Causal explanations and {XAI}.
\newblock In \emph{1st Conference on Causal Learning and Reasoning, CLeaR 2022, Sequoia Conference Center, Eureka, CA, USA, 11-13 April, 2022}, volume 177 of \emph{Proceedings of Machine Learning Research}, pages 90--109. {PMLR}, 2022.

\bibitem[Chockler and Halpern(2004)]{CH04}
H.~Chockler and J.~Y. Halpern.
\newblock Responsibility and blame: {A} structural-model approach.
\newblock \emph{J. Artif. Intell. Res.}, 22:\penalty0 93--115, 2004.

\bibitem[Chockler and Halpern(2024)]{CH24}
H.~Chockler and J.~Y. Halpern.
\newblock Explaining image classifiers, 2024.

\bibitem[Chockler et~al.(2024)Chockler, Kelly, Kroening, and Sun]{chockler2024}
H.~Chockler, D.~A. Kelly, D.~Kroening, and Y.~Sun.
\newblock Causal explanations for image classifiers, 2024.
\newblock URL \url{https://arxiv.org/abs/2411.08875}.

\bibitem[Darwiche and Hirth(2023)]{DH23}
A.~Darwiche and A.~Hirth.
\newblock On the (complete) reasons behind decisions.
\newblock \emph{J. Log. Lang. Inf.}, 32\penalty0 (1):\penalty0 63--88, 2023.

\bibitem[Eiter and Lukasiewicz(2006)]{EL06}
T.~Eiter and T.~Lukasiewicz.
\newblock Causes and explanations in the structural-model approach: Tractable cases.
\newblock \emph{Artificial Intelligence}, 170:\penalty0 542--580, 2006.

\bibitem[Glymour and Wimberly(2007)]{GW07}
C.~Glymour and F.~Wimberly.
\newblock Actual causes and thought experiments.
\newblock In J.~Campbell, M.~O'Rourke, and H.~Silverstein, editors, \emph{Causation and Explanation}, pages 43--67. MIT Press, Cambridge, MA, 2007.

\bibitem[Hall(2007)]{Hall07}
N.~Hall.
\newblock Structural equations and causation.
\newblock \emph{Philosophical Studies}, 132:\penalty0 109--136, 2007.

\bibitem[Halpern(2019)]{Hal19}
J.~Y. Halpern.
\newblock \emph{Actual Causality}.
\newblock The MIT Press, 2019.

\bibitem[Halpern and Pearl(2005)]{HP01b}
J.~Y. Halpern and J.~Pearl.
\newblock Causes and explanations: a structural-model approach. {P}art {I}: causes.
\newblock \emph{British Journal for Philosophy of Science}, 56\penalty0 (4):\penalty0 843--887, 2005.

\bibitem[Hitchcock(2001)]{hitchcock:99}
C.~Hitchcock.
\newblock The intransitivity of causation revealed in equations and graphs.
\newblock \emph{Journal of Philosophy}, XCVIII\penalty0 (6):\penalty0 273--299, 2001.

\bibitem[Hitchcock(2007)]{Hitchcock07}
C.~Hitchcock.
\newblock Prevention, preemption, and the principle of sufficient reason.
\newblock \emph{Philosophical Review}, 116:\penalty0 495--532, 2007.

\bibitem[Ibrahim et~al.(2019)Ibrahim, Rehwald, and Pretschner]{IRP19}
A.~Ibrahim, S.~Rehwald, and A.~Pretschner.
\newblock Efficiently checking actual causality with {SAT} solving.
\newblock \emph{CoRR}, abs/1904.13101, 2019.
\newblock URL \url{http://arxiv.org/abs/1904.13101}.

\bibitem[Ignatiev et~al.(2019)Ignatiev, Narodytska, and Marques{-}Silva]{INMS19}
A.~Ignatiev, N.~Narodytska, and J.~Marques{-}Silva.
\newblock Abduction-based explanations for machine learning models.
\newblock In \emph{The Thirty-Third {AAAI} Conference on Artificial Intelligence, {AAAI}}, pages 1511--1519. {AAAI} Press, 2019.

\bibitem[Kokhlikyan et~al.(2020)Kokhlikyan, Miglani, Martin, Wang, Alsallakh, Reynolds, Melnikov, Kliushkina, Araya, Yan, and Reblitz-Richardson]{captum}
N.~Kokhlikyan, V.~Miglani, M.~Martin, E.~Wang, B.~Alsallakh, J.~Reynolds, A.~Melnikov, N.~Kliushkina, C.~Araya, S.~Yan, and O.~Reblitz-Richardson.
\newblock Captum: A unified and generic model interpretability library for pytorch, 2020.
\newblock URL \url{https://arxiv.org/abs/2009.07896}.

\bibitem[Kullback and Leibler(1951)]{kldivergence51}
S.~Kullback and R.~A. Leibler.
\newblock {On Information and Sufficiency}.
\newblock \emph{The Annals of Mathematical Statistics}, 22\penalty0 (1):\penalty0 79 -- 86, 1951.
\newblock \doi{10.1214/aoms/1177729694}.
\newblock URL \url{https://doi.org/10.1214/aoms/1177729694}.

\bibitem[Lin(1991)]{JS91}
J.~Lin.
\newblock Divergence measures based on the shannon entropy.
\newblock \emph{IEEE Transactions on Information Theory}, 37:\penalty0 145--151, 1991.
\newblock \doi{10.1109/18.61115}.

\bibitem[Lundberg and Lee(2017)]{lundberg2017unified}
S.~M. Lundberg and S.-I. Lee.
\newblock A unified approach to interpreting model predictions.
\newblock In \emph{Advances in Neural Information Processing Systems (Neur{IPS})}, volume~30, pages 4765--4774, 2017.

\bibitem[Marques{-}Silva and Ignatiev(2022)]{MSI22}
J.~Marques{-}Silva and A.~Ignatiev.
\newblock Delivering trustworthy {AI} through formal {XAI}.
\newblock In \emph{Thirty-Sixth {AAAI} Conference on Artificial Intelligence, {AAAI}}, pages 12342--12350. {AAAI} Press, 2022.

\bibitem[Nam et~al.(2020)Nam, Gur, Choi, Wolf, and Lee]{nam2020relative}
W.-J. Nam, S.~Gur, J.~Choi, L.~Wolf, and S.-W. Lee.
\newblock Relative attributing propagation: Interpreting the comparative contributions of individual units in deep neural networks.
\newblock In \emph{AAAI Conference on Artificial Intelligence}, volume~34, pages 2501--2508, 2020.

\bibitem[Petsiuk et~al.(2018)Petsiuk, Das, and Saenko]{petsiuk2018rise}
V.~Petsiuk, A.~Das, and K.~Saenko.
\newblock {RISE:} randomized input sampling for explanation of black-box models.
\newblock In \emph{British Machine Vision Conference ({BMVC})}. {BMVA} Press, 2018.

\bibitem[Priest(2008)]{K3}
G.~Priest.
\newblock \emph{An Introduction to Non-Classical Logic: From If to Is}.
\newblock Cambridge University Press, 2 edition, 2008.

\bibitem[Ribeiro et~al.(2016)Ribeiro, Singh, and Guestrin]{lime}
M.~T. Ribeiro, S.~Singh, and C.~Guestrin.
\newblock ``{W}hy should {I} trust you?'' {E}xplaining the predictions of any classifier.
\newblock In \emph{Knowledge Discovery and Data Mining (KDD)}, pages 1135--1144. {ACM}, 2016.

\bibitem[Ribeiro et~al.(2018)Ribeiro, Singh, and Guestrin]{RSG18}
M.~T. Ribeiro, S.~Singh, and C.~Guestrin.
\newblock Anchors: High-precision model-agnostic explanations.
\newblock In \emph{Proceedings of the Thirty-Second {AAAI} Conference on Artificial Intelligence, (AAAI-18)}, pages 1527--1535. {AAAI} Press, 2018.

\bibitem[Selvaraju et~al.(2017)Selvaraju, Cogswell, Das, Vedantam, Parikh, and Batra]{CAM}
R.~R. Selvaraju, M.~Cogswell, A.~Das, R.~Vedantam, D.~Parikh, and D.~Batra.
\newblock Grad-{CAM}: Visual explanations from deep networks via gradient-based localization.
\newblock In \emph{International Conference on Computer Vision (ICCV)}, pages 618--626. IEEE, 2017.

\bibitem[Shitole et~al.(2021)Shitole, Li, Kahng, Tadepalli, and Fern]{SLKTF21}
V.~Shitole, F.~Li, M.~Kahng, P.~Tadepalli, and A.~Fern.
\newblock One explanation is not enough: Structured attention graphs for image classification.
\newblock In \emph{Neural Information Processing Systems (NeurIPS)}, pages 11352--11363, 2021.

\bibitem[Shrikumar et~al.(2017)Shrikumar, Greenside, and Kundaje]{shrikumar2017learning}
A.~Shrikumar, P.~Greenside, and A.~Kundaje.
\newblock Learning important features through propagating activation differences.
\newblock In \emph{International Conference on Machine Learning (ICML)}, volume~70, pages 3145--3153. {PMLR}, 2017.

\bibitem[Springenberg et~al.(2015)Springenberg, Dosovitskiy, Brox, and Riedmiller]{springenberg2015striving}
J.~T. Springenberg, A.~Dosovitskiy, T.~Brox, and M.~A. Riedmiller.
\newblock Striving for simplicity: The all convolutional net.
\newblock In \emph{ICLR (Workshop Track)}, 2015.
\newblock URL \url{http://arxiv.org/abs/1412.6806}.

\bibitem[Sundararajan et~al.(2017)Sundararajan, Taly, and Yan]{sundararajan2017axiomatic}
M.~Sundararajan, A.~Taly, and Q.~Yan.
\newblock Axiomatic attribution for deep networks.
\newblock In \emph{International Conference on Machine Learning}, pages 3319--3328. PMLR, 2017.

\bibitem[Tritscher et~al.(2020)Tritscher, Ring, Schlr, Hettinger, and Hotho]{EPOC2020}
J.~Tritscher, M.~Ring, D.~Schlr, L.~Hettinger, and A.~Hotho.
\newblock Evaluation of post-hoc xai approaches through synthetic tabular data.
\newblock In \emph{Foundations of Intelligent Systems}, pages 422--430. Springer International Publishing, 2020.

\bibitem[Weslake(2015)]{Weslake11}
B.~Weslake.
\newblock A partial theory of actual causation.
\newblock \emph{British Journal for the Philosophy of Science}, 2015.
\newblock To appear.

\bibitem[Woodward(2003)]{Woodward03}
J.~Woodward.
\newblock \emph{Making Things Happen: A Theory of Causal Explanation}.
\newblock Oxford University Press, Oxford, U.K., 2003.

\bibitem[Yeh et~al.(2019)Yeh, Hsieh, Suggala, Inouye, and Ravikumar]{chih2019infidelity}
C.-K. Yeh, C.-Y. Hsieh, A.~Suggala, D.~Inouye, and P.~Ravikumar.
\newblock On the (in)fidelity and sensitivity of explanations.
\newblock 12 2019.

\end{thebibliography}
